# A Knowledge Acquisition Tool for Bayesian-Network Troubleshooters

**Claus Skaanning**
Hewlett-Packard
Customer Support R&D
Fredrik Bajers Vej 7C
DK-9220 Aalborg SØ
Denmark

## Abstract

This paper describes a domain-specific knowledge acquisition tool for intelligent automated troubleshooters based on Bayesian networks. No Bayesian network knowledge is required to use the tool, and troubleshooting information can be specified as natural and intuitive as possible. Probabilities can be specified in the direction that is most natural to the domain expert. Thus, the knowledge acquisition efficiently removes the traditional knowledge acquisition bottleneck of Bayesian networks.

## 1 INTRODUCTION

This paper describes one of the first of a new type of tools that are starting to appear : domain-specific knowledge acquisition tools for Bayesian networks. It has been acknowledged by many experts and non-experts that one of the major drawbacks of Bayesian networks is the knowledge acquisition bottleneck. For large and complex domains it is a cumbersome task to construct the Bayesian network structure and collect all the required parameter values. On top of that, many of the parameters have to be specified in a manner that is highly unintuitive to the domain expert.

The paper documents the results of a collaboration[1] between the Decision Support Systems group of Aalborg University and Customer Support Research & Development, Hewlett-Packard Company. The goal of this collaboration was to develop intelligent troubleshooting systems for Hewlett-Packard printers using Bayesian networks. The project ran in two phases, (i) provide a proof of concept that efficient troubleshooters based on Bayesian networks can be constructed, (ii) devise means of efficient construction of troubleshooters for domain experts without Bayesian network knowledge. The result of the first phase was the troubleshooting methodology described in the companion paper by Jensen, Skaanning and Kjærulff (2000), and the result of the second phase was the knowledge acquisition tool for Bayesian-network troubleshooters that is described in this paper.

An intelligent troubleshooter guides the customer through a fast sequence of troubleshooting steps (*actions or questions*) attempting to resolve the problem. Currently, the printer industry spends millions of dollars a year on customer support. The majority of these expenses are consumed by call agents providing help over the phone, and support agents sent out to solve problems at the customer site. If an intelligent automated troubleshooter is provided, e.g., over the web, many customers' problems can be resolved without human assistance. If the troubleshooter is unable to find a solution to a customer problem, all the information gathered so far can be transferred to a support agent who will continue the troubleshooting. Given all previously gathered information, the support agent will be able to save much time by skipping steps already performed.

Printer systems consist of many complex components : the application the user is printing from, the printer driver, the network connection, the printer server, the printer itself, etc. It is a very complex task to construct a Bayesian-network troubleshooter for such a large system. Constructing the underlying Bayesian network with traditional graphical tools would be an extremely time-consuming process and would require domain experts trained in Bayesian network theory. These requirements are almost always too restrictive. In order to make Bayesian-network troubleshooters a viable alternative to case-based systems, decision trees, rule-based systems, etc., it is crucial to remove this knowledge acquisition bottleneck.

This paper describes a knowledge acquisition tool that does this. Though the knowledge acquisition tool was developed for printing systems, it is generic and can be used for any application where the basic underlying assumptions are fulfilled. The most crucial is the *single-fault assumption* that requires at most one fault in the system. This is a sound assumption in many scenarios, e.g., printer systems, and it allows enormous simplifications in both the algorithms for deciding the best sequence of steps, and the knowledge acquisition process.

The tool will be referred to as the BATS[2] Author in this paper.

---

[1] The name of the project is SACSO : Systems for Automated Customer Support Operations

[2] Bayesian Automated Troubleshooting System

The knowledge acquisition tool implements the following fundamental ideas:

- the domain expert needs no knowledge of Bayesian networks. All terms are expressed in a manner that is intuitive to the domain expert. All Bayesian network structure is created implicitly by the tool.

- the domain expert can specify probabilities in the most natural / intuitive manner. Domain experts are allowed to specify probabilities in the non-causal direction if they prefer.

- the domain expert specifies potential causes and troubleshooting steps which map to underlying Bayesian network structures. These structures are combined into a Bayesian network for the troubleshooter in a sound manner.

- the tool allows maximum reuse of building blocks consisting of causes and troubleshooting steps. Logical units or components of the domain can be represented as *modules* in a *library* that can be used when constructing troubleshooting models for new error conditions.

The troubleshooting methodology, i.e., how steps are selected and how information is represented in the Bayesian network is covered in more detail in (Skaanning, Jensen and Kjærulff, 2000) and will only be described briefly here.

## 2 TROUBLESHOOTING

Assume that we want to troubleshoot a malfunctioning device with $n$ possible underlying causes represented by the variables $F_1,...,F_n$ ($F$ for fault). In the printing system application, components could for instance be the printer driver, the spooler, etc. Assume that we have defined repair actions $A_1, ..., A_k$, that have the potential to solve the problem, and that each repair action $A_i$ has a probability $P_i = P(A_i = yes \mid e)$ of solving the problem given current evidence, and a cost $C_i$. The cost may be combined from several *cost factors* such as the time it takes to carry out the action, money required to buy requisites, etc. We assume further that we have questions, $Q_1, ..., Q_m$, that can be asked to supply information about the error condition. The questions also have costs, $C_1, ..., C_m$.

In (Skaanning, Jensen and Kjærulff, 2000) it is described how to find good troubleshooting sequences of actions and questions so this will not be discussed further here.

### 2.1 NAIVE BAYES MODEL

The error conditions in the domain are modeled as separate Bayesian networks, one for each error condition. Thus, one separate Bayesian network is created for each error condition. These Bayesian network models will in the following be referred to as error condition models.

When an error condition model has been constructed, the underlying Bayesian network can be read and executed by the accompanying tool, the BATS Troubleshooter.

It is possible to have the models represented as separate Bayesian networks because of the underlying assumption that the user will always be able to identify the correct error condition that he is experiencing. In reality this assumption does not always hold. The current solution to situations where error conditions cannot be easily distinguished, is to represent them in a single large model. This large model can then contain questions that with some uncertainty attempt to determine the error condition.

Each of these error condition models includes a *cause indicator variable* $I$ that defines the probability distribution over the causes of the error condition. The causes are modeled as the states of this variable. All actions and questions that can be posed in the troubleshooting process are represented as children of the cause indicator variable. An example is shown in Figure 1. The benefit of this *naïve Bayes* structure is that all observations (actions and questions) are independent given the causes. This can be exploited in the algorithms for finding the best next action or question as shown in (Skaanning, Jensen and Kjærulff, 2000).

In the following it will be described how causes, actions and questions are represented and how the required information is acquired from domain experts.

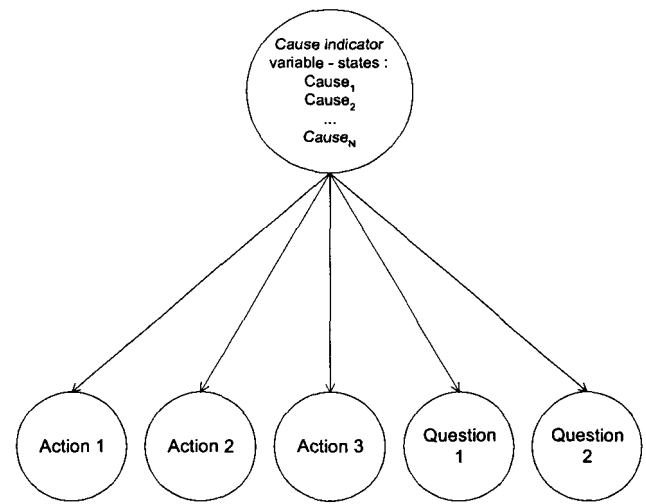

Figure 1 An example of the very simple Bayes structure used for troubleshooters.

## 3 KNOWLEDGE ACQUISITION TOOL

The knowledge acquisition tool implements the basic knowledge acquisition process that was developed during the SACSO project. It is illustrated in Figure 2.

Causes and steps are matched by considering for each action which causes it can potentially solve and for each question which causes it is associated with. During this process actions that do not solve any causes and causes that are not solved by any action are often discovered, requiring the domain expert to go back and add the missing information.

The knowledge acquisition tool allows the user to construct a library of commonly used modules. If such a library exists, new models can be constructed much faster by reusing existing modules with already specified causes and steps. All the stages of Figure 2 except stage #5 can then be skipped for these causes and steps. The prior probabilities of causes will depend on the specific error condition, so these can not be pre-specified in the library module.

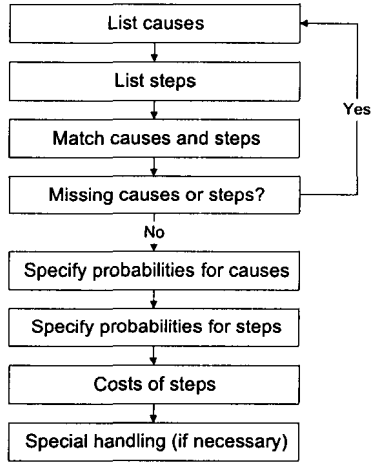

**Figure 2  The knowledge acquisition process.**

In the following sections, the basic building blocks, causes, actions and questions will be described and it will be explained how the knowledge acquisition process is supported by the tool. Finally, it will be described how the building blocks are combined into models or library modules.

### 3.1 CAUSES AND CAUSE TREES

In the knowledge acquisition phase the causes are organized in a tree such that the root of the tree corresponds with the problem-defining node, i.e., the node indicating whether or not the problem is present. The children of the root node are causes or components that, if present or malfunctioning, cause the problem to be present. The children of causes or components are subcauses or causes that, if present, cause the presence of the parent cause, etc. When the final troubleshooter is constructed, the cause tree is collapsed to the cause indicator variable such that each leaf cause corresponds to a state.

An example of a cause tree taken from the printer domain is "Light print"[3] with causes "Toner cartridge", "Paper", and "Settings". "Toner cartridge" has subcauses "Defective toner cartridge" and "Empty toner cartridge".

---

[3] When the user gets output from the printer that is too light.

"Settings" has subcauses "Economode set in application", "Economode set in printer driver", etc.

The causes can always be organized into a tree due to the single-fault assumption. If only a single fault is assumed it is not possible to have subcauses that can cause more than one higher level cause simultaneously. If there is a subcause that can cause more than one higher level cause, then we must have that this subcause can be represented as two mutually exclusive components where each of them causes one of the higher level causes. In this case, the subcause might as well be represented as two independent subcauses each affecting its respective parent cause - so the tree structure is maintained. Thus, due to the single-fault assumption, loops are not possible and so we have a tree of causes. At most one cause at each level of the cause tree can be present.

A small example of such a cause tree is given in Figure 3. In this simple example, we have a problem with two possible causes, and one of these has two possible subcauses. A cause represents an event that if present causes the presence of the parent cause, i.e., the probability of *Problem* given $Cause_1$ in Figure 3 is one, etc.

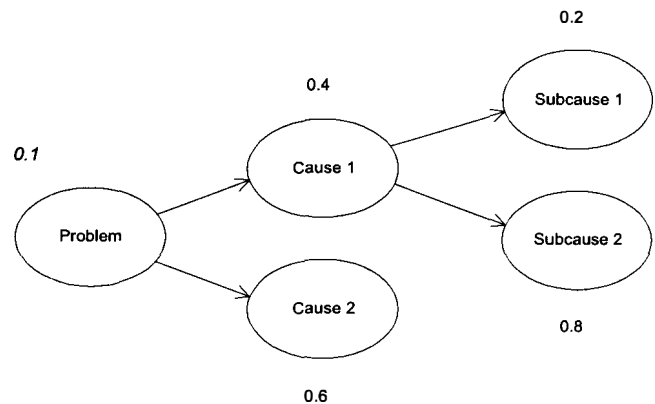

**Figure 3.  A simple Bayesian network with an example probability assignment.**

One cause tree is constructed for each error condition model. The cause tree is constructed by the domain experts. In SACSO, the domain experts were experienced troubleshooters of printer systems, so they had detailed knowledge of causes and subcauses for each error condition.

#### 3.1.1 Knowledge Acquisition of Cause Probabilities

Probabilities for causes are acquired based on the cause tree in the opposite of the causal direction, i.e., the domain experts specify probabilities for causes conditional on the presence of their parent cause. For the example in Figure 3, domain experts have to specify $P(Cause_1 \mid Problem)$, $P(Cause_2 \mid Problem)$, $P(Subcause_1 \mid Cause_1)$ and $P(Subcause_2 \mid Cause_2)$. An example specification of these probabilities can be seen in Figure 3.

There are many advantages to eliciting the probabilities in this manner. The SACSO domain experts were trained at troubleshooting printer system problems, thus they were used to consider a set of causes conditional on the presence of the problem, and in many cases a set of subcauses conditional on the presence of the parent cause when all other causes have been ruled out. Some related work on the benefits on non-causal assessment can be seen in (Schachter and Heckerman, 1987).

With traditional methods, the domain experts assign marginal probabilities for the leaf causes, e.g., $P(Cause_2)$, $P(Subcause_1)$ and $P(Subcause_2)$ in Figure 3. When there are many leaf causes, the probabilities of many of these must be small and so eliciting them becomes harder. When probabilities of leaf causes are assessed assuming the presence of their parent cause, the domain expert only has to consider a small set of causes (those that can cause the parent cause). He must then assess probabilities for this small set of causes such that they sum to 1 which is easier than assessing the unconditional probabilities.

### 3.1.2 Representation in Cause Indicator Variable

The probabilities elicited for the cause tree have to be transformed into a flat probability distribution that can be represented in the cause indicator variable. The cause indicator variable has a state for each leaf cause and the probability of the state is calculated as the probability of the leaf cause given its parent cause multiplied with the product of all the conditional probabilities of its descendants given their parent, i.e., if leaf cause $F$ has parent $F_1$, $F_1$ has parent $F_2$, ..., $F_{k-1}$ has parent $F_k$, and $F_k$ is the root, then

$$P(I = F) = P(F | F_1) \times P(F_k) \prod_{i = 1,..., k-1} P(F_i | F_{i+1})$$

It is not a problem that non-leaf causes are not directly represented as states in the cause indicator variable, as the probabilities of these aggregated causes can be found as the sum of all descendant leaf causes.

One of the underlying assumptions is that the system is only used when there is a problem, thus it is not necessary to have a state representing no problem.

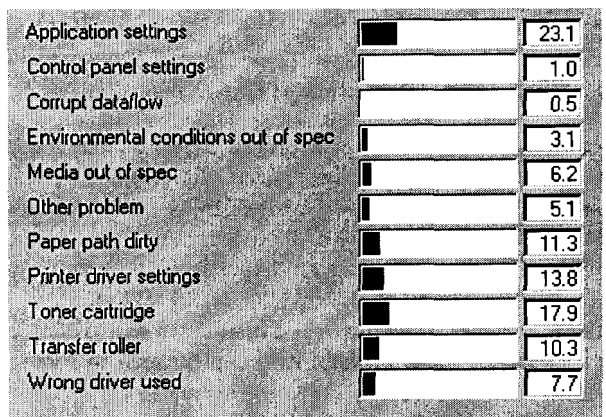

Figure 4 An example of probability assignment for a set of causes in the KA tool.

### 3.1.3 Causes and BATS Author

Probability assignment of causes is simple in the BATS Author. First, a set of causes with the same parent in the cause tree is selected. Then, the user specifies probabilities for the causes conditional on the parent cause and error condition. If the causes in question are on the top level, i.e., they have no parent causes, then the probabilities are specified conditional only on the error condition. The user can use graphical sliders as shown in Figure 4. Figure 4 shows a list of causes of the error condition "Light print" from the printer domain. Other cause information such as the name and an explanation can be specified with a cause editor.

## 3.2 ACTIONS

Actions are troubleshooting steps that when carried out by the user can potentially solve the problem. There are two types of actions, *repair actions* and *test actions*. Repair actions (e.g., reseat the parallel cable) can solve the problem and thus end the troubleshooting process whereas test actions (e.g., try another parallel cable) change the configuration of the system to test whether the problem goes away. A test action thus only provides information that can be used later in the troubleshooting process and the troubleshooting process continues no matter the answer.

The knowledge acquisition and representation of the two types of actions are exactly the same. However, they are treated differently in the algorithms for finding the best next step described in (Skaanning, Jensen and Kjærulff, 2000).

Here, actions are not associated with a single cause as in (Breese and Heckerman, 1996). In practice, enforcing a one-to-one relationship between causes and actions poses much too severe restrictions on the domain experts when building troubleshooters, as many actions naturally affect more than one cause and many causes can be solved by multiple actions.

### 3.2.1 Knowledge acquisition

The knowledge acquisition for actions consists of three steps, (i) listing the causes that the action can solve, (ii) eliciting probabilities that the action solves these causes, (iii) eliciting the cost factors of the action.

If an action solves more than one cause, it is sufficient to elicit one probability for each cause that it solves, i.e., the probability of the action solving the problem assuming that the cause is the actual underlying cause. Due to the single-fault assumption, it is not necessary to consider combinations of the causes.

The assessment of the probability that an action solves a cause has been split into three pieces, (i) $P(A=yes | F, correct, requisites)$, the probability of the action solving the problem assuming that the cause $F$ is present, the

action is performed correctly and all requisites are in perfect order, (ii) *P(correct)*, the probability that the action is performed correctly, and (iii) *P(requisites)*, the probability that all requisites are in working order. The three probabilities are then combined into one :

$$P(A = \text{yes}|F) = P(A = \text{yes}|F, \text{correct}, \text{requisites}) \times P(\text{correct}) \times P(\text{requisites}) \quad (1)$$

This probability elicitation is much easier for the domain expert if it is split into three pieces as they do then not have to simultaneously balance several factors in their minds. In the above calculation the events *correct* and *requisites* are independent of the presence of *f* and they are independent of each other.

The following cost factors are elicited for troubleshooting steps, (i) *time*, (ii) *risk* - of breaking something else, (iii) *money* - required to carry out the step, (iv) *insult*[4] - potential insult to user if this step is suggested. Time is elicited in minutes and money in dollars. Risk and insult, however, are specified on a scale from 0 to 4. When eliciting the cost factors, the domain experts have to average over the user population intended for the troubleshooter.

The cost factors are combined linearly to form the overall cost of the step :

$$C = \alpha T + \beta R + \gamma M + \delta I$$

The weights can be determined by having the domain experts perform many preference elicitations where they select the cost combination they prefer from a list of choices.

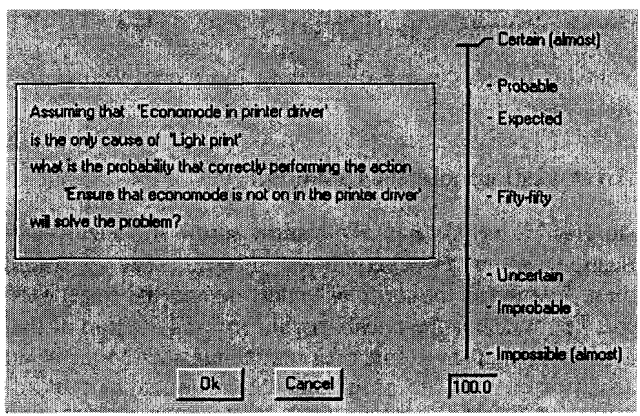

**Figure 5  Eliciting the probability of an action solving a cause.**

### 3.2.2 Representation

Actions are represented as children of the cause indicator variable. Thus, in the probability table for the action, we will need to specify the probability of the action solving the problem given each possible cause. For causes that the action cannot solve, this probability will be 0. For causes that the action can solve, the probability will be $P(A=\text{yes} \mid f)$ as found above. It is not necessary to provide a probability that the action will solve the problem if there is no problem, as one of the underlying assumptions of the Bayesian network is that it is only used when there is a problem present.

### 3.2.3 Actions and BATS Author

The BATS Author has an action editor for specifying actions in an error condition model. This action editor allows the domain expert to specify the causes that are solved by the action. Again, a graphical slider can be used to set the probability of the action solving a cause, as seen in Figure 5 where we have a typical elicitation - an action that solves a cause with absolute certainty. As the action is assumed to be performed correctly, these elicitations are mostly very simple and rapidly performed. The graphical slider in Figure 5 was borrowed from (van der Gaag et al., 1999).

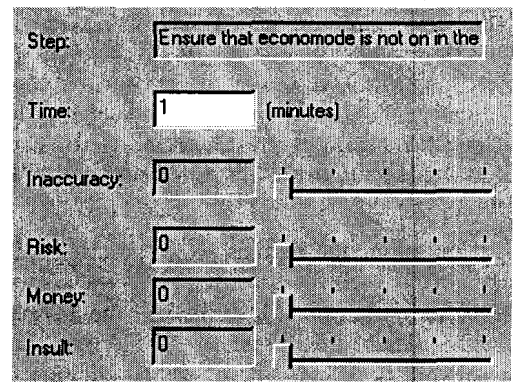

**Figure 6  Cost editor**

The cost editor is depicted in Figure 6. The various cost factors can be specified with graphical sliders. Time is specified in minutes but with the other factors a five level scale is used (0=none, low, medium, high, very high).

Strictly, the inaccuracy factor should not have been placed in the cost editor as it is not a cost factor. The inaccuracy factor is used for specifying the uncertainty involved with the action, i.e., how much the user's result is trusted. This factor is converted into *P(correct)* used in Eq. (1) above.

## 3.3 QUESTIONS

Questions are troubleshooting steps that gathers information that can be used in the troubleshooting process. There are basically two types of questions, symptom questions and general questions. *Symptom* questions concern symptoms or effects of the problem, and *general* questions concern something that could have caused or created the problem.

---

[4] In printer systems, steps such as "Check whether the printer is turned on" or "Check whether the parallel cable is plugged in" can be insulting to experienced users.

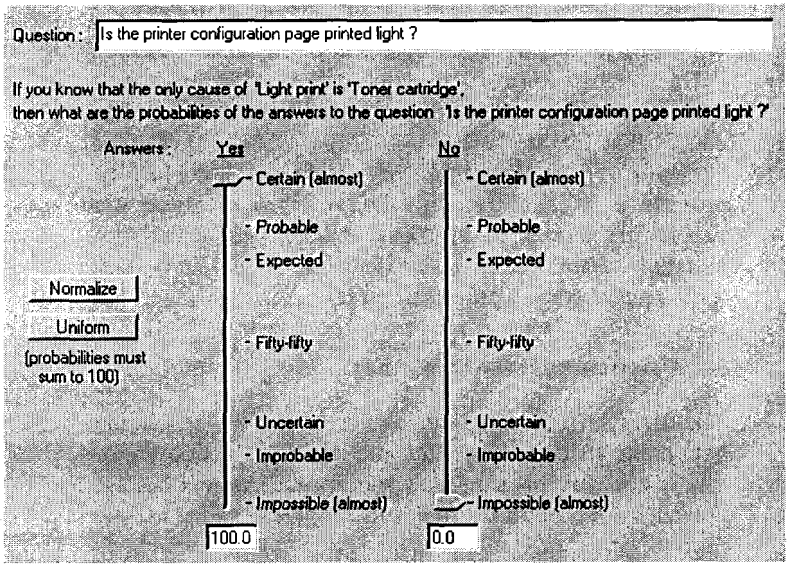

**Figure 7 Probability elicitation for a question that concerns a symptom.**

For both types of questions, the domain expert must first list causes that are associated with the question.

For symptom questions, the domain experts must elicit the probabilities of the question answers given each associated cause, and probabilities of the answers given that none of the associated causes are present. Questions are represented as children of the cause indicator variable. Symptom questions are easy to represent as the elicited probabilities can be used directly in the probability table of the node.

For general questions the domain experts must elicit the probabilities of the associated causes given the answer to the question, and the prior probabilities of the answers to the question. Thus, the direction of the arcs in the Bayesian network is reversed here. When eliciting these probabilities, the following equation must be maintained:

$$P(F) = \sum_{Q=s} P(F \mid Q=s) P(Q=s) \quad (2)$$

Thus, the probabilities that must be specified here are very much interdependent. The BATS Author provides an interactive screen with sliders for all related probabilities in Eq. (2). When one slider is dragged, all affected sliders are updated correspondingly.

General questions are also represented as children of the cause indicator variable in the Bayesian network. To represent them this way, however, we need to reverse the probabilities from $P(F \mid Q)$ to $P(Q \mid F)$. This can be done by applying Bayes' formula:

$$P(Q \mid F) = \frac{P(F \mid Q) P(Q)}{P(F)}.$$

Due to the single-fault assumption and assuming that questions are independent of each other given the causes, it can be shown that the probabilities can always be reversed using Bayes' formula. This is simple to see in the situation where one question affects one cause. When one question affects a set of causes it gets more complicated.

Assuming a question $Q$ that affects causes $F_1, ..., F_k$, the user has then elicited probabilities, $P(F_1 \mid Q), ..., P(F_k \mid Q)$, and $P(Q)$ and we need to find $P(Q \mid F_1, ..., F_k)$. Expressed in terms of the cause indicator variable $I$, the user has elicited probabilities $P(I=F_1 \mid Q), ..., P(I=F_k \mid Q)$, and $P(Q)$ and we need to find $P(Q \mid I=F_1), ..., P(Q \mid I=F_k)$, and $P(Q \mid I \neq F_1, ..., I \neq F_k)$. We find these as follows:

$$P(Q \mid I = F_i) = \frac{P(I = F_i \mid Q) P(Q)}{P(I = F_i)},$$

$$P(Q \mid I \neq F_1, ..., I \neq F_k) = \frac{P(I \neq F_1, ..., I \neq F_k \mid Q) P(Q)}{P(I \neq F_1, ..., I \neq F_k)}.$$

And

$$P(I \neq F_1, ..., I \neq F_k \mid Q) = 1 - \sum_{1 \leq i \leq k} P(I = F_i \mid Q).$$

The single-fault assumption enables us to reverse the arcs as simply as shown above. Interestingly, this allows the domain expert to specify the conditional probabilities of multiple causes given a question one cause at a time without considering the other causes and it is certain that the multiple effects of the question on the causes will be implemented as intended.

In the general case of multiple questions affecting a cause, we assume independence between the questions and perform the reversal for each question without taking others into consideration. If there is a dependence between two questions, the domain expert can represent them as a composite question with BATS Author.

Costs for questions are elicited in the same way as for actions, see Figure 6.

### 3.3.1 Questions and BATS Author

The BATS Author has a question editor for specifying questions in an error condition model. The question editor first helps the domain expert decide the question's type, i.e., whether the question concerns a symptom of the causes, or whether it concerns something that could have caused or created the problem. Determining the type of the question allows the BATS Author to provide the domain expert with the most intuitive and natural way of eliciting the probabilities, i.e., as the probability of the question conditional on the causes, or as the probabilities of causes conditional on the question.

Typically, the domain expert is not familiar with conditional probabilities and is not able to choose the best way to elicit the probabilities, so by determining the type of the question the BATS Author leads directly to the best method.

There are questions that are not easily categorized but it is our experience that for these questions it usually does not matter which way the probabilities are elicited.

For very simple types of questions that either eliminate or identify causes based on the answer, the BATS Author provides short cuts to allow the domain expert to directly specify the eliminated or identified causes. Of course, depending on the question type, he will still have to either specify prior probabilities for the question answers, or probabilities of non-eliminated and non-identified causes conditional on the question answers.

**P(Question | Causes)**

The BATS Author provides graphical sliders for editing the probabilities of question answers conditional on a specific cause being present, see Figure 7.

**P(Causes | Question)**

For specifying the probabilities of causes conditional on question answers, the BATS Author provides the domain expert with two levels of precision. In Figure 8 and Figure 9 are shown the approximate specification method. In Figure 8 is shown an overview of the causes that are associated with the question and arrows indicate how the probability of the cause is modified for each answer to the question. Figure 9 shows the choices available.

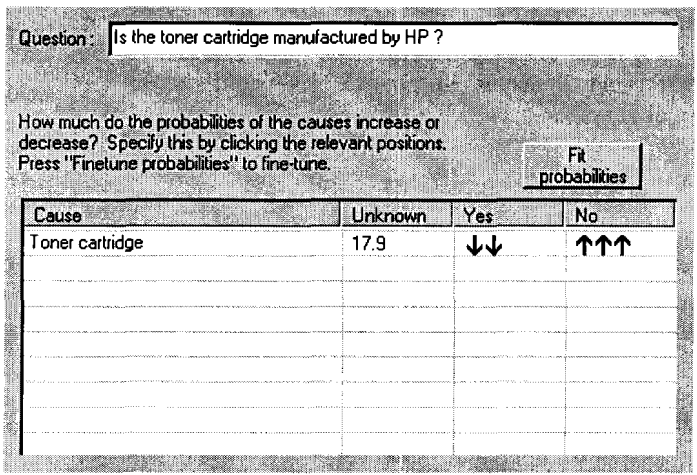

**Figure 8 Probability elicitation for a question that concerns something that could have caused the problem.**

As mentioned earlier, the probabilities specified here always have to satisfy Eq. (2), so the domain expert's selection of arrows may not necessarily be consistent with other probabilities. The button labeled "Fit probabilities" in Figure 8 starts an algorithm that attempts to satisfy as many of the domain expert's wishes as possible. The algorithm simply runs through the causes one at a time, attempting to satisfy the wishes for that cause. If the wishes cannot be satisfied, it is attempted to satisfy them partly, e.g., change three up arrows to two, etc. The assumption is that it is better to satisfy two wishes partly than satisfy one wish completely and not satisfy another one.

Often, the approximate specification in Figures 8 and 9 is sufficient - and often the domain expert is not capable of assessing the probabilities with any higher degree of accuracy anyway. However, in some situations more precision and accuracy is required. For these situations the BATS Author provides a highly complex set of sliders for all the probabilities in Eq. (2) where any slider can be set as wanted, and all affected sliders will be adjusted accordingly.

### 3.4 Combining the Pieces

In the previous sections it was explained how the three pieces, causes, actions and questions, could be specified with the BATS Author. Causes are constructed as a tree which is later collapsed to a single cause indicator variable in the Bayesian network. Actions and questions are represented as children of this cause indicator variable. The simplicity of this Bayesian network allows for very efficient algorithms for finding the best sequence of steps.

The BATS Author allows the domain expert to construct a new error condition model from scratch by following the KA process laid out in Figure 2, or, if a library of modules exists, appropriate modules can be used to quickly populate a new error condition model.

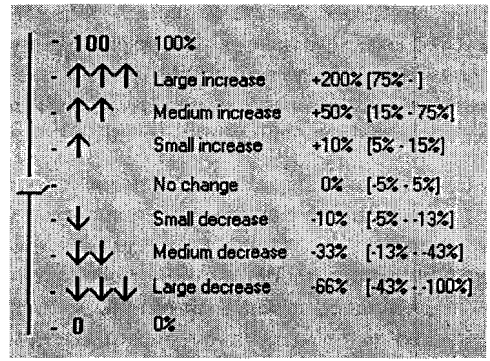

**Figure 9 Probability elicitation for questions.**

The benefits of allowing reuse can be illustrated with examples from the printer domain. For example, the toner cartridge component can be a cause of many different error conditions, most of them related with print quality. It is our experience that with a well populated library of modules, the development time for new error conditions can be cut at least in half, as 70-80% of the causes can be directly copied from the library.

Thus, the BATS Author allows for a great deal of reuse between models. Further, the tool maintains consistency between shared modules such that when information changes in a library module this is mirrored in all the error condition models where this module was used.

The BATS Author also provides excellent search and replace facilities that makes maintenance and migration highly efficient.

## 4 DISCUSSION AND CONCLUSIONS

In this paper we have presented a novel knowledge acquisition tool for construction of troubleshooters. The BATS Author is quite general and allows construction of Bayesian-network troubleshooters for any device or system that fulfills the basic assumption of a single fault.

### 4.1 Experience in the use of the BATS Author

It is very easy to learn to use the tool. All 6-7 domain experts associated with the SACSO project learned to use older versions of the BATS Author in just a few days, and none of these had any prior knowledge of Bayesian networks. It is our experience that the basic structure and features of the current tool can be learned in a single day, perhaps just a few hours - and the more advanced features can be learned in less than a week. When the SACSO project started, all troubleshooter models had to be constructed by hand in an expert system shell like Hugin which could only be performed by someone knowledgeable with Bayesian networks. The BATS Author does not require any knowledge of Bayesian networks, so it has provided a huge leap forward.

Of course, it is also beneficial to use the BATS Author for people that has experience with Bayesian networks, if they want to construct troubleshooters. The BATS Author only requests the minimal necessary parameters and constructs the rest of the probability tables itself. The ability to reuse model structure is an obvious advantage for anyone with or without Bayesian knowledge.

It is hard to provide evidence of the efficiency of the BATS Author as there isn't anything to compare it with. Our current experience is that a set of error condition models for one printer model (70-80 models) can be constructed and validated in one man-month. In fact this was accomplished in one month by one of our domain experts with an early version of the tool. If a set of models for the previous printer model exists, it may be possible to migrate to the new printer in just a few days. The BATS Author provides several mechanisms that aid the migration such as search / replace in all text.

So far approximately 200 error condition models have been constructed with the BATS Author for four different printer models.

One of the major benefits of the BATS Author is that it allows extensive reuse of model components. Out of the 200 error condition models constructed so far, approximately, 30-40 were constructed from scratch, and the remainder were constructed from pieces of the first, sometimes almost a direct copy with only a few changes in the text descriptions.

We claim that the BATS Author is general, i.e., it can be used for any device or system fulfilling the single-fault assumption. Though we do not have much experience to base this claim on, the BATS Author has been successfully used to construct models in domain such as networking and automobiles. For these domains no extra functionality was required.

### 4.2 Related work

Knowledge acquisition tools for constructing troubleshooters exist for decision trees and case based systems; however, we have been unable to find reference of any such tool within the area of Bayesian networks. Indeed, we have been unable to find reference of any knowledge acquisition tool within the area of Bayesian networks that does not require Bayesian expertise. This leads us to believe that the BATS Author is in fact one of the first tools for knowledge acquisition based on Bayesian networks.

It seems likely that it will not be the last, though. There must be many application areas where knowledge acquisition can be simplified and tuned such that only the absolutely necessary information is required to construct networks, and such that this information can be provided in the manner most natural to the available domain experts.

Heckerman (1991) provided a general approach to knowledge acquisition for diagnostic applications with the so-called similarity networks. A Macintosh based similarity network editor named "SimNet" was implemented based on this approach. The approach is more general than the one we have proposed, but on the other hand it seems to provide less guidance to the domain experts.

### 4.3 Future research

There are a few problems with the approach taken in the BATS Author. One of the assumptions is that all error conditions can be represented as separate Bayesian networks. This assumption does not always hold in practice. Print quality problems in the printer domain, for instance, represent an instance of this problem. It is quite difficult for the user to distinguish between spots, bands and stripes on the paper; however, identifying the correct error condition is crucial in getting the optimal troubleshooting sequence.

The solution we have adopted is to represent error conditions that are hard to distinguish in a single large model. This model can then contain questions that with some uncertainty attempts to determine the correct error condition. However, even if it is not possible to completely determine the error condition, actions can be selected based on the probabilities of the error conditions contained in the model.

We will also be investigating a more flexible solution to this problem in the future. It should be possible to handle situations where a user enters an incorrect model, and from the provided answers determine another model to switch to. The other model should then be populated with the answers provided in the previous model. Further, it should be possible to construct a *super-model* that attempts to determine the correct error condition by asking a sequence of questions. When sufficient information has been gathered, this super-model can then start the most likely error condition. If, after some time, it seems unlikely that this was the correct error condition, the super-model takes over again and routes the user to the next most likely model, possibly after asking new questions.

One other potential problem with our approach is the method for specifying cause probabilities. The domain experts have to specify these without assuming any prior knowledge. In fact, domain experts are often used to consider these probabilities in very specific scenarios, knowing for instance the operating system, certain symptoms, etc. In these circumstances it may be more efficient and easier for the domain experts to provide the cause probabilities conditional on the answers to certain questions, and then determine the unconditional cause probabilities from this.

When troubleshooting, modifications of the device under test are part of the process. As a result, a troubleshooting action can potentially invalidate previous observations. Means for representing these dependencies were also implemented in the BATS Author, making it possible to specify, e.g., that when action A is performed, question Q will be fixed in the specified state.

When eliciting probabilities for causes, the device is assumed to be in a faulty state, i.e., the domain expert does not have to specify P(problem). The Bayesian network is thus created without a "Normal" state. This is an acceptable representation if the Bayesian network is only used when the device is malfunctioning. We do foresee a need to specify P(problem) to be able to perform better model selection and better handling of the situation where the true underlying problem is different from the one handled by the current selected one.

In the current version of the authoring tool it is possible to specify how cost factors are combined, so that different versions of the model could be created for users with different levels of expertise.

The current version of the BATS Author requires the single-fault assumption to be fulfilled. The tool could be modified to allow multiple faults, using, e.g., an assumption of independent causes (Srinivas, 1995). The algorithms for finding the best next step (Skaanning, Jensen and Kjærulff, 2000) rests heavily on the single-fault assumption, so they would have to be modified.

## 5 ACKNOWLEDGEMENTS

The author would like to thank those on the SACSO team who have helped making the BATS Author a success during the last two years of hard work : Finn V. Jensen, Uffe Kjærulff, Janice Bogorad, Lasse Rostrup-Jensen, Paul Pelletier, Lynn Parker, and Brian Kristiansen. Further, the author would like to thank everyone in the Decision Support Systems Group of Aalborg University for providing an inspiring environment.